# Explainable Artificial Intelligence Model for Evaluating Shear Strength Parameters of Municipal Solid Waste Across Diverse Compositional Profiles

Parichat Suknark[1,3], Sompote Youwai[2,*], Tipok Kitkobsin[2], Sirintornthep Towprayoon[1,3], Chart Chiemchaisri[4] and Komsilp Wangyao[1,3]

[1]The Joint Graduate School of Energy and Environment,
King Mongkut's University of Technology Thonburi, Bangkok 10140, Thailand
[2]AI Research Group, Department of Civil Engineering, Faculty of Engineering,
King Mongkut's University of Technology Thonburi, Bangkok 10140, Thailand
[3]Center of Excellence on Energy Technology and Environment (CEE), PERDO, Ministry of Higher Education, Science, Research and Innovation (MHESI), Bangkok 10140, Thailand
[4]Department of Environmental Engineering, Faculty of Engineering, Kasetsart University,
 Bangkok 10900, Thailand

*Corresponding author, Email: sompote.you@kmutt.ac.th, https://orcid.org/0009-0002-9878-8504

## Abstract

Accurate prediction of shear strength parameters in Municipal Solid Waste (MSW) remains a critical challenge in geotechnical engineering due to the heterogeneous nature of waste materials and their temporal evolution through degradation processes. This paper presents a novel explainable artificial intelligence (XAI) framework for evaluating cohesion and friction angle across diverse MSW compositional profiles. The proposed model integrates a multi-layer perceptron architecture with SHAP (SHapley Additive exPlanations) analysis to provide transparent insights into how specific waste components influence strength characteristics. Training data encompassed large-scale direct shear tests across various waste compositions and degradation states. The model demonstrated superior predictive accuracy compared to traditional gradient boosting methods, achieving mean absolute percentage errors of 7.42% and 14.96% for friction angle and cohesion predictions, respectively. Through SHAP analysis, the study revealed that fibrous materials and particle size distribution were primary drivers of shear strength variation, with food waste and plastics showing significant but non-linear effects. The model's explainability component successfully quantified these relationships, enabling evidence-based recommendations for waste management practices. This research bridges the gap between advanced machine learning and geotechnical engineering practice, offering a reliable tool for rapid assessment of MSW mechanical properties while maintaining interpretability for engineering decision-making.

# 1. Introduction

The mechanical stability of Municipal Solid Waste (MSW) landfills constitutes a fundamental challenge in geotechnical engineering, particularly given the accelerating rates of waste generation associated with global urbanization trends [1–3]. Landfill stability analysis represents a critical design parameter that significantly influences both the engineering design and operational protocols. The structural integrity of these geotechnical systems is predominantly governed by the waste material's shear strength parameters, specifically cohesion (c) and internal friction angle ($\phi$) [4, 5]. The characteristic heterogeneity of MSW compositions, in conjunction with temporal variations induced by progressive biodegradation processes [3, 6, 7], presents substantial challenges in the accurate determination of these parameters through conventional analytical methodologies. While recent developments in Artificial Intelligence (AI) have demonstrated promising capabilities in modeling complex geotechnical systems [8, 9], the application of these computational approaches to MSW characterization remains insufficiently investigated. The primary challenge extends beyond the development of predictive algorithms to encompass the necessity for model interpretability and validation in practical engineering contexts. This investigation addresses this knowledge gap through the development of an explainable AI framework optimized for the evaluation of shear strength parameters across heterogeneous MSW compositional matrices.

Contemporary investigations into MSW shear strength characteristics have elucidated substantial variability in mechanical properties, with the Mohr-Coulomb failure criterion ($\tau = c + \sigma \tan(\phi)$) establishing the theoretical foundation for stability analysis [3, 7]. The shear strength characteristics of MSW exhibit primary dependencies on waste composition, unit weight, and degradation kinetics. Empirical investigations have documented substantial variations in cohesion (1.17-40.17 kPa) and friction angles (21.51°-50.60°) across diverse experimental conditions [6, 10]. While conventional laboratory methodologies provide critical insights, the inherent heterogeneity of waste matrices presents persistent challenges for behavioral prediction. Experimental investigations by Pulat and Yukselen-Aksoy [11] have demonstrated that elevated paper content in synthetic MSW correlates positively with cohesion while exhibiting an inverse relationship with friction angles, whereas Bray et al. [12] and Chen et al. [13] documented that increased plastic content generally attenuates both parameters. Furthermore, Dixon and Jones [14] established that waste degradation processes typically manifest in diminished cohesion values accompanied by enhanced friction angles, highlighting the temporal evolution of MSW mechanical properties. These complex interdependencies, coupled with thermal sensitivity analyses demonstrating inverse correlations between temperature and cohesion, underscore the necessity for advanced analytical methodologies capable of addressing multivariate relationships and parameter estimation uncertainties in MSW characterization [4, 5]. The accurate prediction of shear strength variations across heterogeneous waste compositions remains a significant challenge in contemporary geotechnical research.

Artificial Intelligence (AI) has demonstrated extensive applications in civil engineering, particularly in predictive modeling and structural damage detection through computer vision [8, 15]. However, the interpretability of AI-generated results remains challenging, leading to the emergence of Explainable Artificial Intelligence (XAI). XAI addresses the transparency requirements in AI systems by developing methods to elucidate decision-making processes across various sectors, including healthcare and finance [16, 17]. In geotechnical engineering, XAI methodologies have enhanced model interpretability in slope stability analysis [18], liquefaction prediction [19], and tunneling assessments [20]. XAI techniques reveal parameter influences on both local and global predictions, enabling geotechnical engineers to understand model characteristics and confidently apply AI systems in practical applications, despite limited expertise in deep learning architectures. This research introduces a novel approach that leverages advanced machine learning techniques while maintaining transparency through explainable AI methodologies. By incorporating tools such as SHAP (SHapley Additive exPlanations), the proposed framework not only predicts shear strength parameters but also provides clear insights into how different waste components contribute to these predictions. This explainability component is crucial for building trust among engineering practitioners and ensuring the practical applicability of the model in real-world scenarios.

The primary objectives of this study are to develop a robust artificial intelligence (AI) model for predicting shear strength parameters of municipal solid waste (MSW) across diverse compositional profiles, implement explainability mechanisms to elucidate the relationship between waste composition and mechanical strength, and validate the model's predictions against empirical data from laboratory tests and documented landfill stability case studies. These objectives collectively address the need for reliable, interpretable methods in MSW characterization. To enhance practical utility, a web-based application hosted in public cloud space was proposed, enabling engineers to access the model and visualize the contributions of individual waste components to shear strength predictions. This tool leverages SHAP (SHapley Additive exPlanations) to quantify parameter influences, thereby fostering transparency and trust in AI-driven decision-making for MSW engineering. The dataset employed for model training was derived from the largest Preksa-controlled open dumpsite in Thailand. Shear strength measurements were obtained via large-scale direct shear tests conducted on waste samples with varying compositions and physical conditions, including unit weight. These experiments systematically explored the influence of waste heterogeneity on mechanical behavior, providing a comprehensive dataset to train and evaluate the AI framework.

The remainder of this paper is structured to systematically address these objectives through a comprehensive review of relevant literature, focusing on both traditional geotechnical approaches to MSW characterization and recent applications of AI in geotechnical engineering. Following this, we detail the methodology, including data collection, model architecture, and the implementation of explainability tools. The subsequent sections present the results and validation studies, discuss the implications for landfill design and waste management practices, and conclude with recommendations for future research directions.

## 2. Model architecture

The model architecture employed in this study was a multi-level perceptron (MLP), a feedforward neural network designed to capture non-linear relationships within the dataset [21]. Given that the municipal solid waste (MSW) data comprised a single time-series dataset with 18 features, the MLP was configured to process input variables representing waste composition, physical properties, and environmental conditions. These features included parameters such as waste type proportions, moisture content, unit weight, and compaction history, which were derived from empirical measurements conducted at the study site. The MLP architecture was optimized through hyperparameter tuning to ensure robust predictive performance while mitigating overfitting risks [21]. The dataset, sourced from the largest Preksa-controlled open dumpsite in Thailand, was preprocessed to normalize feature values and address temporal dependencies inherent in the time-series structure. This preprocessing step ensured compatibility with the MLP's input requirements and enhanced the model's ability to generalize across diverse MSW profiles. The resulting framework not only predicted shear strength parameters with high accuracy but also facilitated interpretability through post-hoc analysis techniques, such as SHAP (SHapley Additive exPlanations), which quantified the contribution of individual features to model predictions. The architecture of model as follows:

-First hidden layer (64 neurons):
$$h_1 = \text{ReLU}(W_1 x + b_1) \tag{1}$$
$$h_1 = \text{Dropout}(h_1, p = 0.2) \tag{2}$$

-Second hidden layer (1000 neurons):
$$h_2 = \text{ReLU}(W_2 h_1 + b_2) \tag{3}$$
$$h_2 = \text{Dropout}(h_2, p = 0.2) \tag{4}$$

-Third hidden layer (200 neurons):
$$h_3 = \text{ReLU}(W_3 h_2 + b_3) \tag{5}$$
$$h_3 = \text{Dropout}(h_3, p = 0.2) \tag{6}$$

-Fourth hidden layer (8 neurons):
$$h_4 = \text{ReLU}(W_4 h_3 + b_4) \tag{7}$$
$$h_4 = \text{Dropout}(h_4, p = 0.2) \tag{8}$$

-Output layer
$$y = W_5 h_4 + b_5 \tag{9}$$

-Matrix dimensions
$$W_1 \in R^{64 \times \text{iptsz}} \tag{10}$$
$$W_2 \in R^{1000 \times 64} \tag{11}$$
$$W_3 \in R^{200 \times 1000} \tag{12}$$
$$W_4 \in R^{8 \times 200} \tag{13}$$
$$W_5 \in R^{1 \times 8} \tag{14}$$

-Complete forward pass equation
$$y = W_5 \, \text{Dropout} \left( \text{ReLU} \left( W_4 \, \text{Dropout} \left( \text{ReLU} \left( W_3 \, \text{Dropout} \left( \text{ReLU} \left( W_2 \, \text{Dropout} \left( \text{ReLU} (W_1 x + b_1) \right) + b_2 \right) \right) + b_3 \right) \right) + b_4 \right) \right) + b_5 \tag{15}$$

Where:

- $W_i$ are weight matrices initialized using Xavier/Glorot uniform initialization
- $b_2$ are bias vectors initialized to zero

- ReLU($x$)=max(0,x)
- Dropout$(x,p)$ randomly sets elements of x to zero with probability $p$=0.2

## 3. Dataset

The Praeksa controlled open dumpsite in Samut Prakan province, Thailand, represents a complex waste management facility, as evidenced by comprehensive aerial orthographic imagery and ground-level documentation [22]. The aerial survey (Fig. 1) reveals a sophisticated operational layout with distinct functional zones: an active disposal area marked in cyan forming a curved perimeter, and an inactive zone marked in yellow along the eastern boundary. Two specific monitoring areas, designated as FW (Fresh Waste) and YW (Young Waste) zones, are strategically positioned within the site for waste behavior assessment. The facility's scale is substantial, spanning approximately 300 meters across, with significant elevation variations ranging from 15 meters below grade to 37 meters above the surrounding ground level, creating a total vertical profile of 52 meters.

The site's operational complexity is further compounded by critical stability challenges, clearly visible in Fig. 2, which highlights the risk of insufficient stability in a very small portion of the controlled dumpsite. The ground-level photograph reveals problematic geotechnical conditions characterized by tension cracks, deformation, and inadequate structural integrity of the waste mass. These issues are exacerbated by the heterogeneous composition of the municipal solid waste, which includes various materials such as plastics and textiles, contributing to unpredictable shear strength characteristics [10, 23]. The visible steep slope angles and vertical faces in the waste mass indicate potential shear failure planes, highlighting the urgent need for enhanced geotechnical monitoring and improved waste placement procedures.

The investigation of shear strength in municipal solid waste through machine learning approaches has become increasingly critical due to the complex and heterogeneous nature of waste materials. As evidenced in Fig. 2, waste components vary significantly, including plastics, textiles, organic matter, paper, and other materials, each with distinct mechanical properties. Traditional geotechnical testing methods often struggle to accurately predict shear strength due to this heterogeneity and the dynamic nature of waste decomposition Machine learning algorithms can process multiple variables simultaneously [24–26], including waste composition percentages, age, degree of decomposition, moisture content, and density, to develop more accurate predictive models for shear strength parameters. This is particularly important because different waste components contribute differently to the overall mechanical behavior - for instance, fibrous materials like textiles and plastics often provide reinforcement effects, while organic matter degradation can lead to significant changes in strength properties over time.

The site's daily operations, processing approximately 3,000 Mg of unsorted municipal waste, have created distinct zones based on waste degradation status [22, 27]. The inactive zone contains mature, stabilized waste at least three years old, designated for refuse-derived fuel (RDF) production, while the active zone comprises partially degraded waste less than one year in age, including fresh deposits. The aerial imagery also reveals essential infrastructure supporting daily operations, including access roads and buffer zones separating the facility from surrounding areas. This highlights key operational practices, emphasizing the importance of effective leachate and gas management, as well as the need for enhanced slope stability measures [28–31]. The documented conditions underscore the critical importance of implementing enhanced waste placement techniques and comprehensive geotechnical monitoring protocols to ensure the long-term stability and safety of this essential waste management facility.

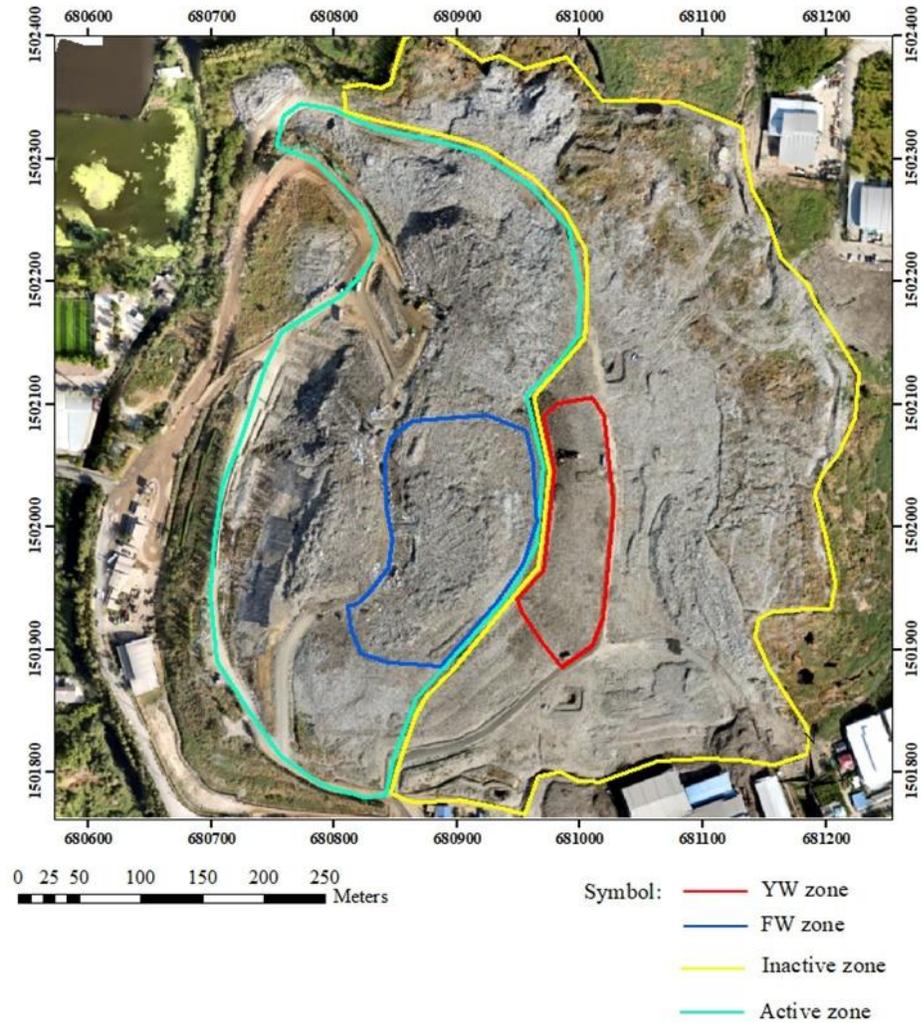

**Fig. 1** Ortho image of the Praeksa controlled dumpsite in Samut Prakan province, Thailand, showing different operational zones.

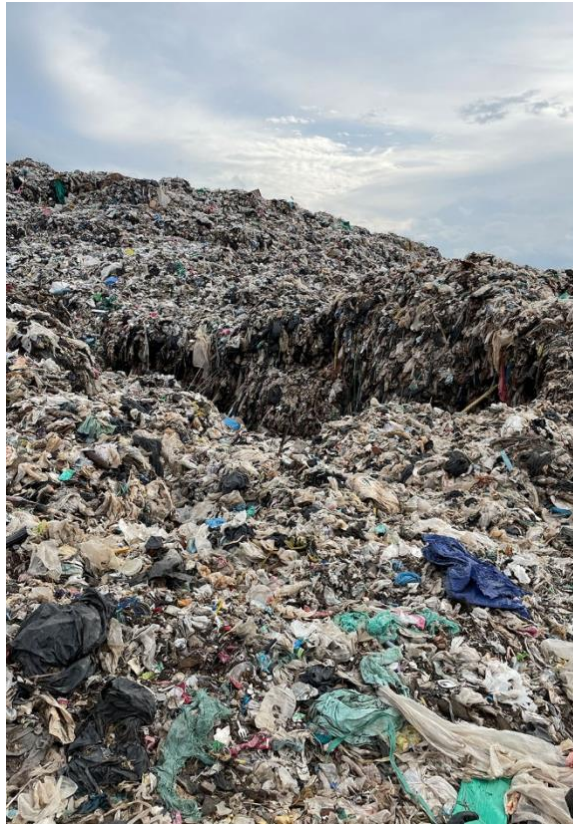

**Fig. 2** A minor area with slight stability concerns

### 3.1 Shear strength parameters measurement using a direct shear test

Direct shear testing was implemented as the primary methodology for shear strength determination based on three critical advantages: (1) accommodation of larger specimens (up to 90 cm) enabling representative sampling of MSW heterogeneity, (2) direct measurement of horizontal displacement under controlled normal stress, and (3) simplified specimen preparation compared to triaxial methods. These advantages specifically address the limitations of triaxial testing, which is constrained by specimen size requirements that may underrepresent material variability [5]. Testing protocols adhered to ASTM D3080 specifications for consolidated drained conditions, ensuring standardization and result reproducibility.

The testing apparatus design, as illustrated in Fig. 3, prioritized structural rigidity and measurement precision. The frame, constructed from 5×5 cm steel tubing (170 cm × 40 cm) (component 9), was dimensioned to minimize deflection under loading, while the lower shear box (components 3, 40×40×10 cm internal dimensions, 2 mm steel plate) and upper shear box (components 4, 40×40×35 cm internal dimensions, 2 mm steel plate) incorporated 5 cm reinforcement flanges to maintain geometric stability during testing. Force application system selection was based on precision control requirements, utilizing a YNT-01 linear actuator (component 1, 300 mm extension, 3,000 N maximum force) with pulse-width modulation control to ensure consistent displacement rates. Measurement instrumentation positioning was optimized for accuracy: the 10 kN YLR-3 load ring (component 5) was mounted 10 cm above the box base to minimize moment effects, while the UNI-T LM 50 laser meter (component 2) was positioned at 30 cm height to ensure stable displacement readings. Normal stress conditions (10.00, 15.00 and 20.00 kPa) were selected to simulate shallow burial depths (1.23 to 4.77 m) based on multiple critical factors: (1) reduced frictional resistance under low confining pressures increases failure susceptibility, (2) enhanced vulnerability to water infiltration affects material properties, (3) exposure to environmental factors including thermal cycling and biochemical degradation modifies waste characteristics, and (4) potential for initiating progressive failure mechanisms influences overall stability [3, 7, 32]. The consolidation criterion (volume change <0.01%, minimum consolidation strain >5.00%) was established based on recent research demonstrating optimal specimen preparation conditions [33].

Shear loading parameters were determined through mechanical considerations: the 20 mm/min displacement rate was selected to maintain drained conditions while minimizing testing duration. Due to the absence of distinct failure planes in several test cases, characteristic of MSW heterogeneous composition [5], the shear strength parameters were consistently evaluated at 20 mm displacement. This standardized displacement criterion was adopted to ensure uniform strength parameter determination across all specimens, regardless of their individual stress-strain behavior. Data acquisition intervals (1 mm) were optimized to provide sufficient resolution for accurate determination of shear stress evolution while maintaining manageable data volumes. This systematic approach to parameter selection enabled reliable derivation of cohesion and friction angle parameters, facilitating comprehensive characterization of MSW shear strength behavior under representative field conditions.

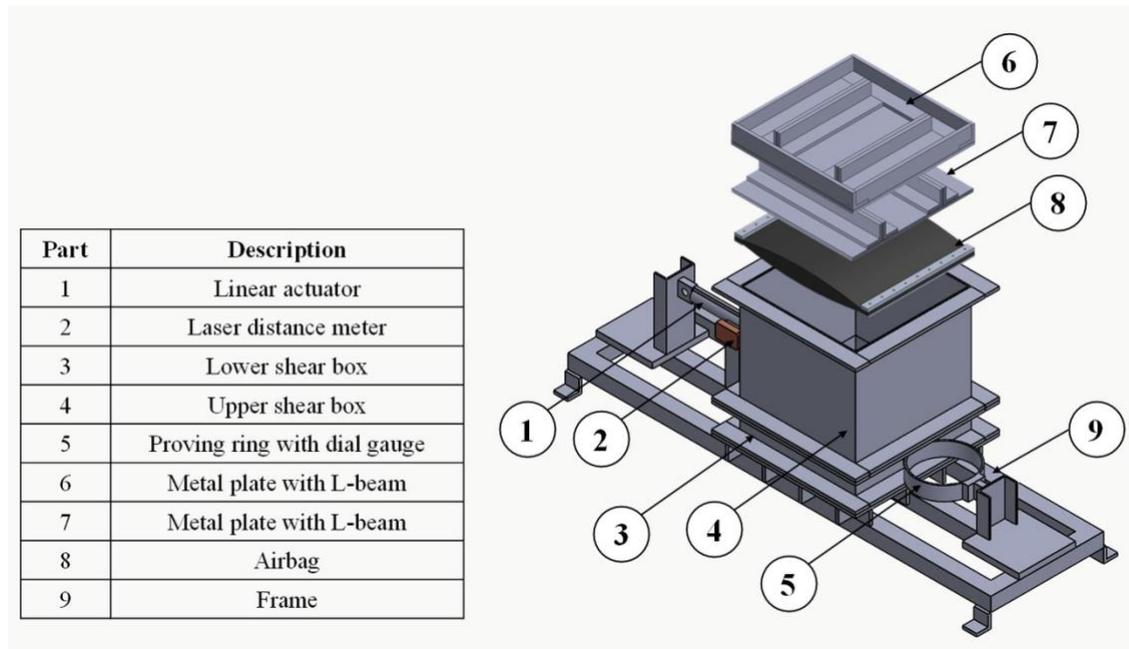

Fig. 3 Illustration of schematic direct shear test procedure used in this study

### 3.2 Data characteristics

The Fig. 4 offers a comprehensive visualization of waste material distributions and physical properties in municipal solid waste management through a series of 15 histogram distributions. The material compositions reveal several distinct patterns: food waste demonstrates a pronounced right-skewed distribution with maximum frequency in the 0.00-0.05 range, while garden waste exhibits an exponential decay-like pattern concentrated within the 0.00-0.02 interval. The minimize proportion of food and garden waste (organic) represents partially degraded waste agree with [31, 34]. Paper and cardboard waste presents an interesting multimodal distribution with multiple distinct peaks, suggesting various sources or types of paper waste streams. Apart from that is the different waste age and moisture content in waste dump resulting in various of biodegradable phase of paper and cardboard, based on this study measure waste proportion by weight [35]. Textiles show a near-normal distribution centered approximately at 0.05, and plastics display a notable bimodal distribution pattern between 0.40 and 0.70. The high plastics proportion, as a significant waste, represents that the decreasing of organic proportion from biodegradation and transform into soil-likes materials or fine fractions [36]. The physical properties of the waste materials provide additional insights into their characteristics. The particle size distributions, categorized into multiple size ranges (10-15 mm, 5-10 mm, 2-5 mm, and <2 mm), exhibit varying degrees of normality, indicating diverse fragmentation patterns in the waste stream. The moisture content, measured in percentage weight per weight (%w/w), shows a right-skewed distribution with a peak in the 0.50-0.60 range, while density measurements in $kN/m^3$ approximate a normal distribution centered around 7.00 to represent the controlled open dump density [34, 37]. These patterns reveal important information about the physical nature of the waste materials and their potential handling characteristics.

The Fig. 5 illustrates a detailed analysis of feature distributions for outlier detection across diverse waste material categories and physical characteristics. The plot encompasses 16 distinct features along the x-axis, spanning from organic materials like food and garden waste to synthetic materials such as plastics and rubber, along with particle size classifications and moisture content measurements. Each boxplot's structure reveals the statistical distribution, with the box representing the interquartile range (IQR) and the green horizontal line indicating the median, while whiskers extend to non-outlier extremes and circles mark statistical outliers. The analysis reveals several significant patterns: plastics demonstrate the highest median value (approximately 0.43) and the largest IQR, indicating substantial variability in plastic content across samples; moisture content (<%wt/wt) shows notable spread with multiple upper-range outliers, suggesting heterogeneous water distribution due to leachate accumulation from the lack of leachate collection system [22]; and most material categories exhibit positively skewed distributions, as evidenced by outlier concentrations above the upper whiskers. Particle size classifications (ranging from 10-15 mm to <2 mm) display relatively consistent distributions with fewer outliers, indicating more uniform size distribution characteristics. The standardized y-axis scale (0.00-0.8) enables direct comparison between features, facilitating quantitative assessment of relative concentrations and variations across waste material categories. This comprehensive visualization effectively highlights the inherent variability and potential anomalies across different waste material features, providing valuable insights for machine learning-based waste classification and sorting systems.

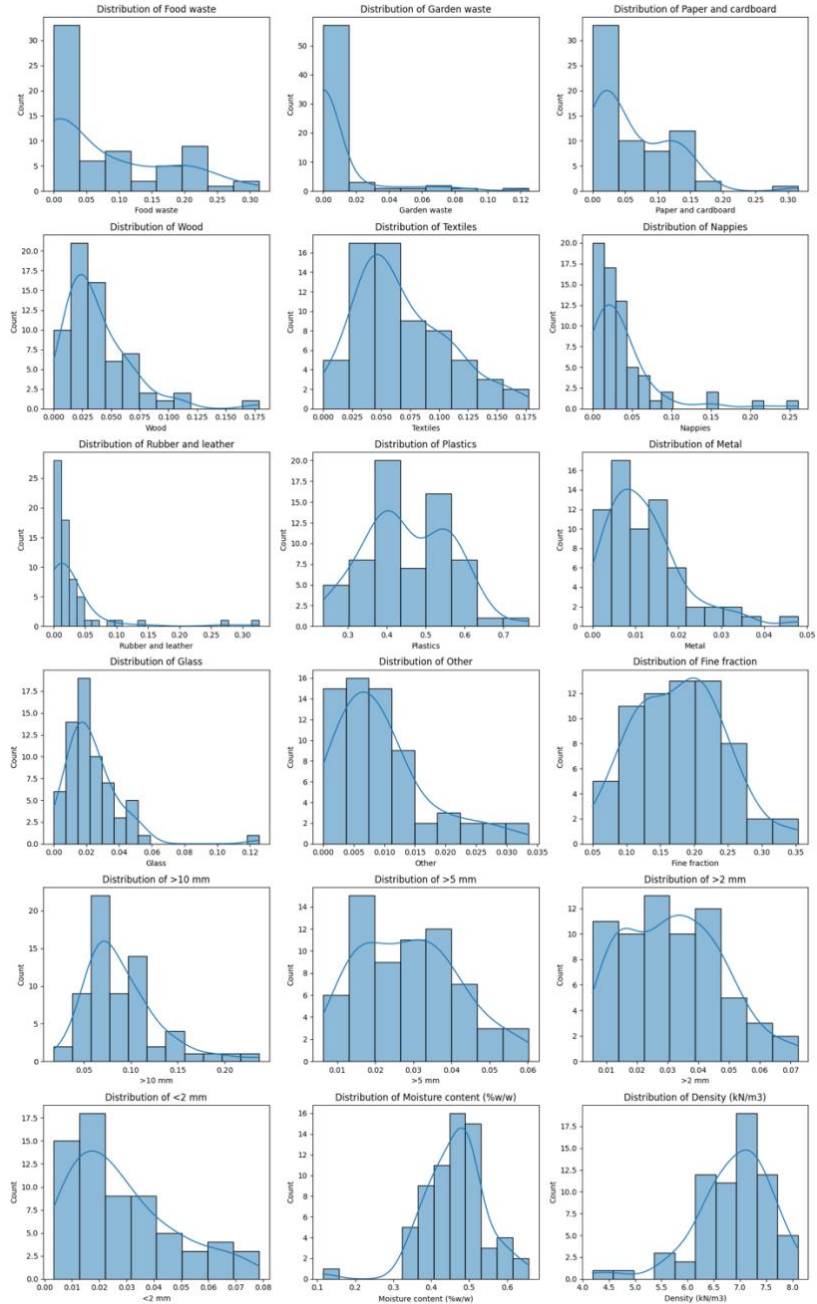

**Fig. 4** The feature distribution of the model

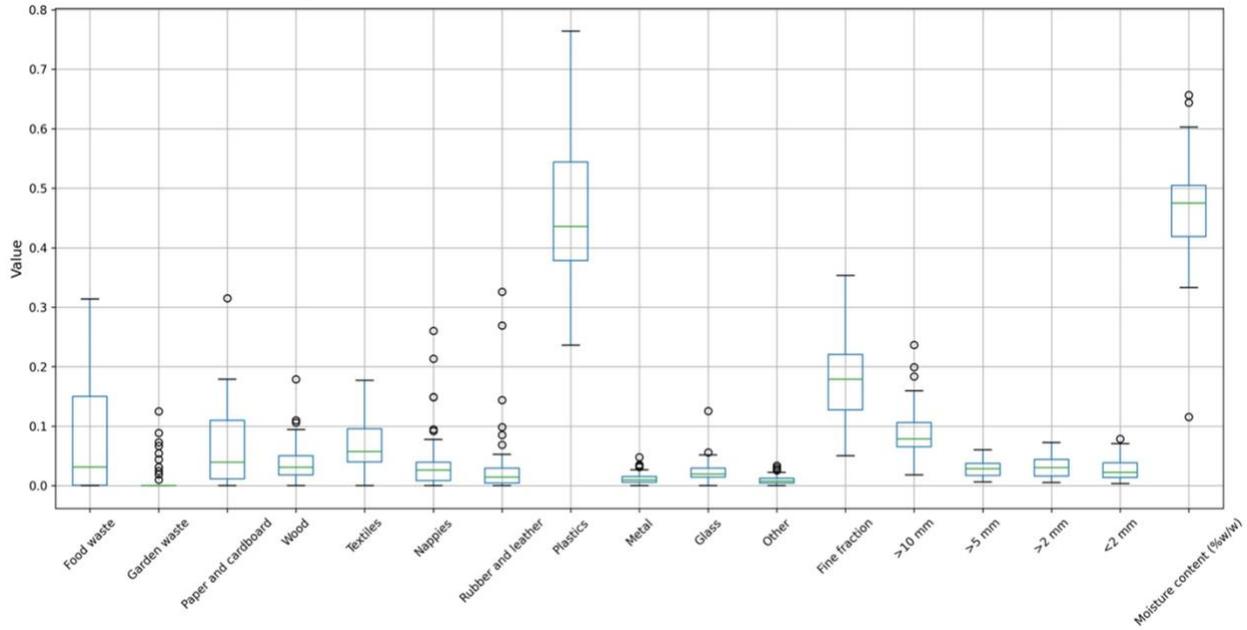

**Fig. 5** The distribution of the feature in the model

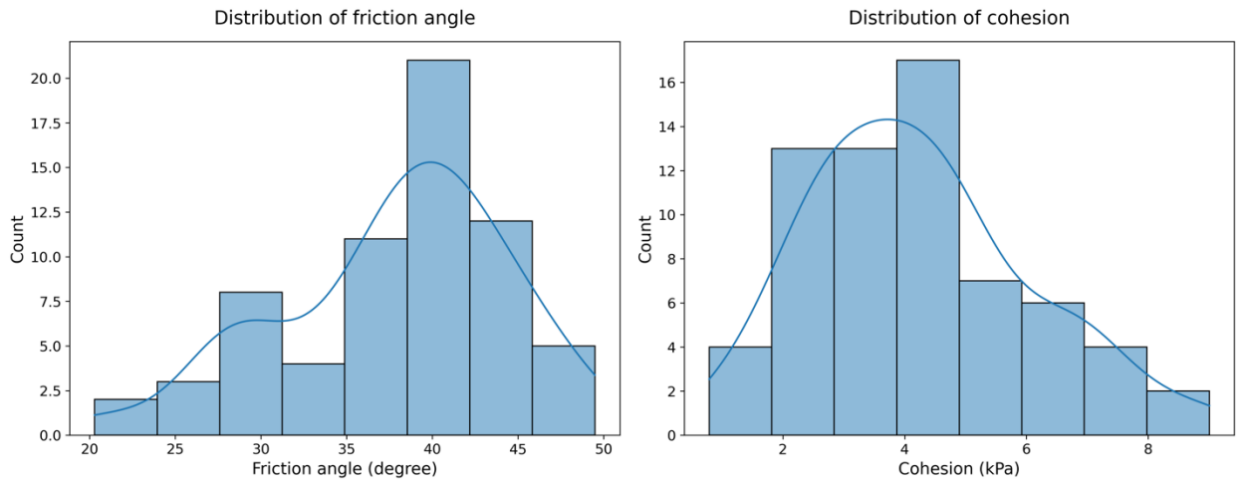

**Fig. 6** The target of prediction distribution

The Fig. 6 presents two histograms depicting the statistical distributions of key geotechnical parameters: friction angle ($\phi$) and cohesion (c). These distributions offer insights into the variability of soil mechanical properties across sampled specimens. The friction angle distribution (left panel) exhibits approximately normal characteristics with a slight positive skewness. The data ranges from 20° to 50°, with the modal class centered at approximately 40-45°. The distribution demonstrates a primary peak frequency of approximately 20 counts, with secondary frequencies observed in the lower ranges, particularly around 30° agree with the typical waste [3, 5, 38]. The superimposed probability density function suggests a reasonable fit to the normal distribution, albeit with some deviation in the tails. The cohesion distribution (right panel) spans from 2 to 9 kPa and displays a more pronounced right-skewed pattern. The peak frequency occurs in the 4-5 kPa range with approximately 17 counts, followed by a gradual decay toward higher cohesion values. The results of cohesion relatively lower than typical waste due to low normal stress to represent shallow waste which is the most expose to the environment [7, 39]. The fitted probability curve indicates a non-symmetric distribution with positive skewness, characteristic of many geotechnical parameters. These distributions are particularly relevant for probabilistic analyses in geotechnical engineering and machine learning applications, where understanding the underlying statistical nature of soil parameters is crucial for model development and uncertainty quantification. The apparent non-normality in both distributions suggests the need for appropriate statistical transformations when implementing these parameters in AI-based predictive models.

### 4. Experiment

The input features X and target variable y (friction angle ϕ and cohesion) were preprocessed using the Min-Max Scaler to linearly transform the data into a standardized range ([0,1]) while preserving the original distribution of the samples. This method was selected to avoid distortions introduced by alternative scaling techniques, such as standardization (which centers data and scales to unit variance, potentially altering relationships) or logarithmic transformation (which non-linearly modifies data structure and may distort skewness or introduce biases for zero/negative values). Both input features and target variables were normalized prior to model training to ensure consistent scaling, mitigate scale-dependent biases in gradient-based optimization algorithms, and retain the statistical properties critical for accurate representation of geotechnical phenomena (e.g., friction angle and cohesion). By maintaining the original distribution, the Min-Max Scaler ensures that the model learns relationships without artificial distortions, which is particularly important for tasks requiring precise physical interpretation.

$$X_{scaled} = \frac{X - X_{min}}{X_{max} - X_{min}} \tag{16}$$

$$y_{scaled} = \frac{y - y_{min}}{y_{max} - y_{min}} \tag{17}$$

The model was trained using the AdamW optimizer, a variant of the Adam algorithm that decouples weight decay from gradient-based updates, thereby enhancing regularization consistency and generalization performance across different architectures. This approach was selected to mitigate overfitting and ensure stable convergence, as AdamW's decoupled weight decay avoids interference with adaptive learning rates, leading to more reliable optimization compared to standard Adam with L2 regularization. The Mean Squared Error (MSE) loss function was employed to quantify prediction errors, as it provides a differentiable, scale-sensitive metric that penalizes large residuals proportionally, making it suitable for regression tasks involving continuous variables like friction angle and cohesion.

$$\text{MSE} = \frac{1}{n} \sum_{i=1}^{n} (y_i - \hat{y}_i)^2 \tag{18}$$

The learning rate ($lr_{epoch}$) was dynamically adjusted using a StepLR scheduler (initial learning rate $lr_0$=0.005, step size=300 epochs, decay factor $\gamma$=0.8), which progressively reduces the learning rate to enable coarse exploration of the parameter space early in training and fine-tuned convergence in later stages, thereby improving optimization stability and preventing overshooting of minima.

$$lr_{epoch} = lr_0 \cdot \gamma^{\left\lfloor \frac{epoch}{stepsize} \right\rfloor} \tag{19}$$

Gradient clipping (clip norm=1.0) was applied to prevent exploding gradients, a common issue in deep learning that can destabilize training by causing abrupt weight updates

$$|\nabla| = \min\left(|\nabla|, clip_{norm}\right) \text{ where } \left(clip_{norm} = 1.0\right) \tag{20}$$

By limiting gradient magnitudes, this technique ensures stable parameter updates, enhances generalization by reducing overfitting risks, and maintains compatibility with activation functions sensitive to large inputs. Together, these strategies were chosen to balance optimization efficiency, numerical stability, and model robustness, ensuring reliable performance for geotechnical prediction tasks.

The model's performance was comprehensively evaluated using three complementary metrics: Mean Absolute Error (MAE), Mean Absolute Percentage Error (MAPE), and the Coefficient of Determination ($R^2$). MAE quantifies the average magnitude of prediction errors in the target variable's native units, providing an interpretable measure of absolute deviation critical for assessing precision in geotechnical predictions. MAPE normalizes errors relative to actual values, offering a percentage-based score to identify systematic over- or underestimation biases and ensure scale-invariant interpretability, particularly in applications where proportional accuracy is prioritized. $R^2$ assesses the proportion of variance in the target variable explained by the model, with values near 1.0 indicating strong explanatory power and effective pattern capture. Together, these metrics balance absolute error magnitude (MAE), relative error impact (MAPE), and global trend alignment ($R^2$), enabling robust identification of both local prediction precision and global model fit. This multi-metric approach ensures alignment with application-specific requirements for reliability and generalizability, safeguarding against over- or underestimation of model performance.

$$\text{MAE} = \frac{1}{n}\sum_{i=1}^{n}|y_i - \hat{y}_i| \tag{21}$$

$$\text{MAPE} = \frac{100\%}{n}\sum_{i=1}^{n}\left|\frac{y_i - \hat{y}_i}{y_i}\right| \tag{22}$$

$$R^2 = 1 - \frac{\sum_{i=1}^{n}(y_i - \hat{y}_i)^2}{\sum_{i=1}^{n}(y_i - \bar{y})^2} \tag{23}$$

The limited dataset size (66 samples) raised concerns regarding the model's generalization capability, prompting the adoption of 10-fold cross-validation to robustly assess performance while maximizing data utilization. Cross-validation is a standard practice for small datasets, as it minimizes bias in performance estimation by iteratively partitioning the data into training (90%) and validation (10%) subsets. The results demonstrated an average Mean Absolute Error (MAE) of 5.1±1.4 degree for friction angle and 0.5±0.3 kPa for cohesion across validation folds. These low MAE values, coupled with narrow standard deviations, indicate consistent predictive accuracy and stability across folds, thereby validating the model's generalization despite the small dataset size. The findings further support the model's applicability for predicting the shear strength of municipal solid waste (MSW), a critical parameter in geotechnical engineering. The use of cross-validation here aligns with established guidelines for evaluating machine learning models on constrained datasets, ensuring reliable performance estimation and mitigating overfitting risks. The low error magnitudes relative to the target variables' scales (e.g., friction angles typically ranging from 20°–40°) further substantiate the model's practical utility for MSW shear strength prediction.

Following cross-validation, a 90:10 train-test split was implemented to evaluate the model's final performance on an independent, unseen dataset. While cross-validation provides robust internal validation by leveraging all data for both training and validation, it inherently lacks an external holdout set to assess generalization to novel samples. The 90:10 split ensures the model is trained on the largest feasible subset (90% of data) while reserving 10% as a strictly independent test set, which is critical for unbiased evaluation of real-world predictive capability. This approach aligns with best practices in machine learning, where cross-validation is used for hyperparameter tuning and model selection, while a separate test set provides a final, objective performance metric. Given the dataset's small size, the 10% test subset strikes a balance between retaining sufficient training data for model learning and ensuring a representative sample for external validation. The test set results serve as the definitive measure of the model's generalization, complementing the cross-validation findings and providing stakeholders with confidence in its applicability to unseen MSW shear strength predictions.

In the realm of hyperparameter optimization for machine learning models, a systematic approach to exploring hyperparameter spaces involves generating all possible value combinations through the Cartesian product of candidate sets. This method ensures thorough coverage of the hyperparameter space, enabling the identification of optimal configurations.

Let $H = \{H_1, H_2, \ldots, H_P\}$ denote a set of hyperparameters, with each $H_p$ having a discrete set of candidate values $S_p = \{s_{p1}, s_{p2}, \ldots, s_{pK_p}\}$ The Cartesian product of these sets:

$$C = S_1 \times S_2 \times \cdots \times S_p \tag{24}$$

produces all possible ordered combinations of values: $s\_\{1i\}, s\_\{2j\}, \ldots, s\_\{Pk\})$ where $s\_\{1i\}, s\_\{2j\}, \ldots, s\_\{Pk\})$. For example, combining learning rates $\{0.001, 0.01, 0.1\}$ and batch sizes $\{16, 32, 64\}$ results in nine unique configurations. Evaluating each combination during grid search systematically identifies the best-performing hyperparameters by ensuring no combination is overlooked.

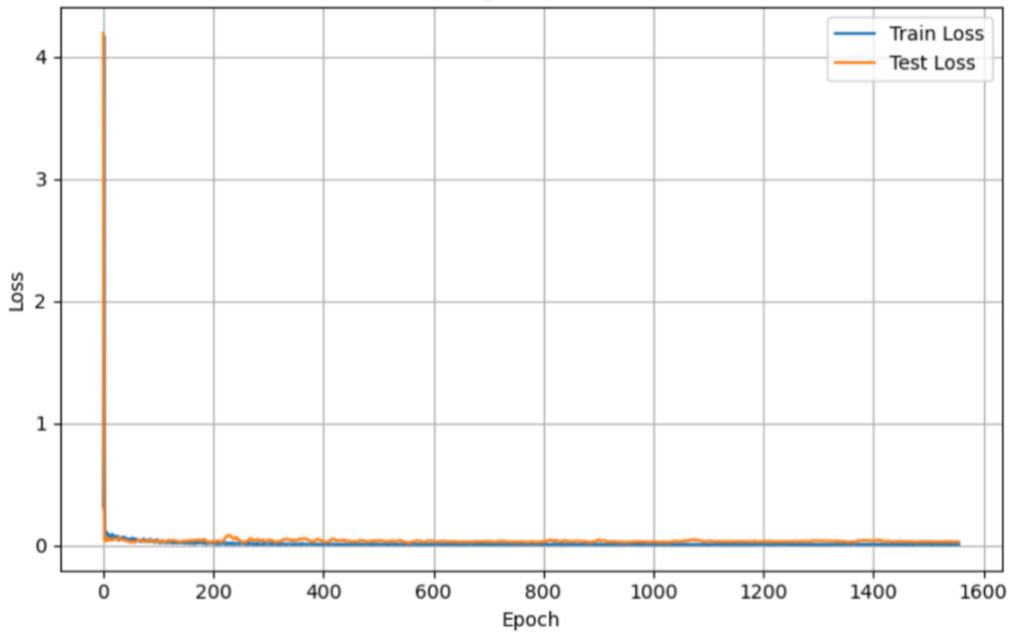

**Fig. 7** Training and test loss curves over epochs for the friction angle prediction model.

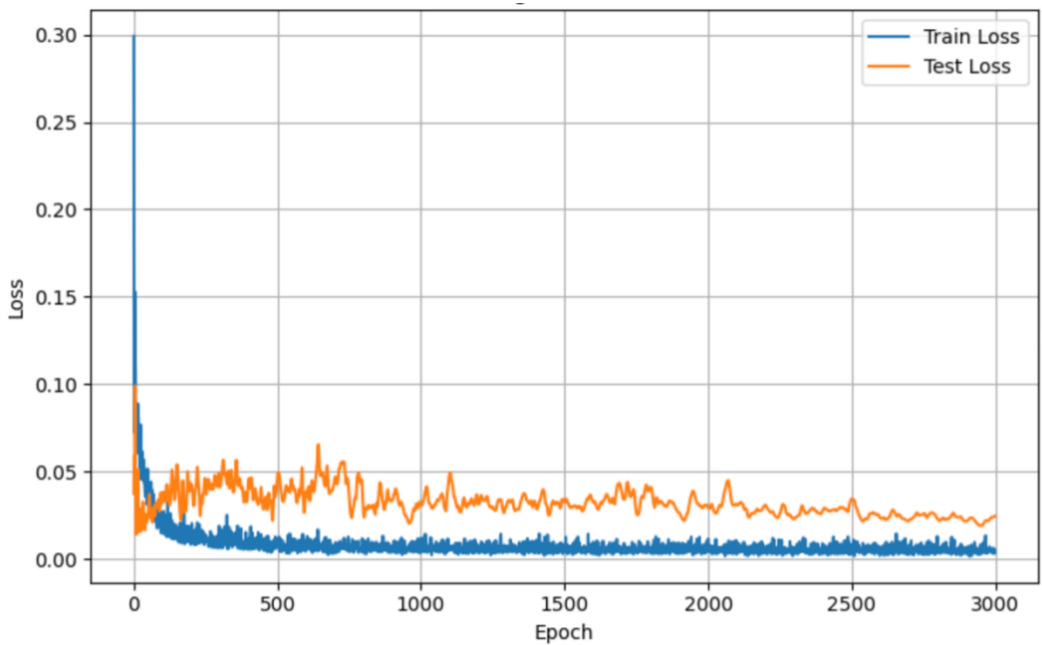

**Fig. 8** Training and test loss curves over epochs for the cohesion prediction

Figs. 7 and 8 illustrate the training and test loss trajectories for the friction angle and cohesion prediction models, respectively, providing critical insights into their learning dynamics and generalization behavior. For the friction angle model (Fig. 3), both training and test losses rapidly converged within the initial 200 epochs, stabilizing at near-zero values with minimal divergence. This tight alignment indicates robust generalization, as the model maintained consistent predictive accuracy on unseen data, likely attributed to effective regularization (AdamW with weight decay) and a well-calibrated model capacity relative to the dataset size. In contrast, the cohesion prediction model (Fig. 4) exhibited a persistent gap between training and test losses, with the latter plateauing at a slightly elevated level (≈0.03–0.05) despite the training loss approaching zero. The fluctuating test loss, particularly during early epochs, suggests mild overfitting, potentially due to weaker feature-target correlations or higher noise sensitivity in the cohesion data. The slight overfitting may also stem from inherent noise or fluctuations in the test data, possibly attributable to friction-induced stick-slip phenomena during experimental measurements, which can introduce variability in cohesion readings at low values. While this noise may contribute to minor discrepancies between training and test performance, the absolute magnitude of the test loss remains within an acceptable range, indicating that the model retains sufficient predictive utility for practical applications.

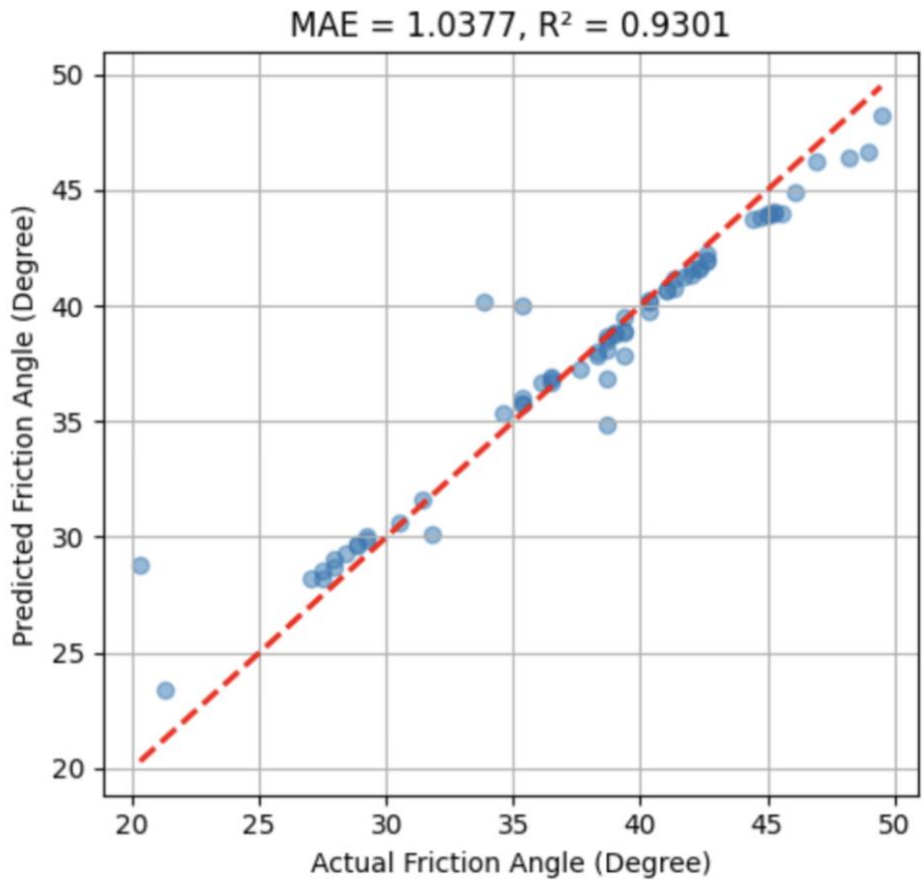

**Fig. 9** Model Performance for Friction Angle Prediction: Predicted vs. Actual Values (degree) (MAE = 1.0377, $R^2$ = 0.9301)

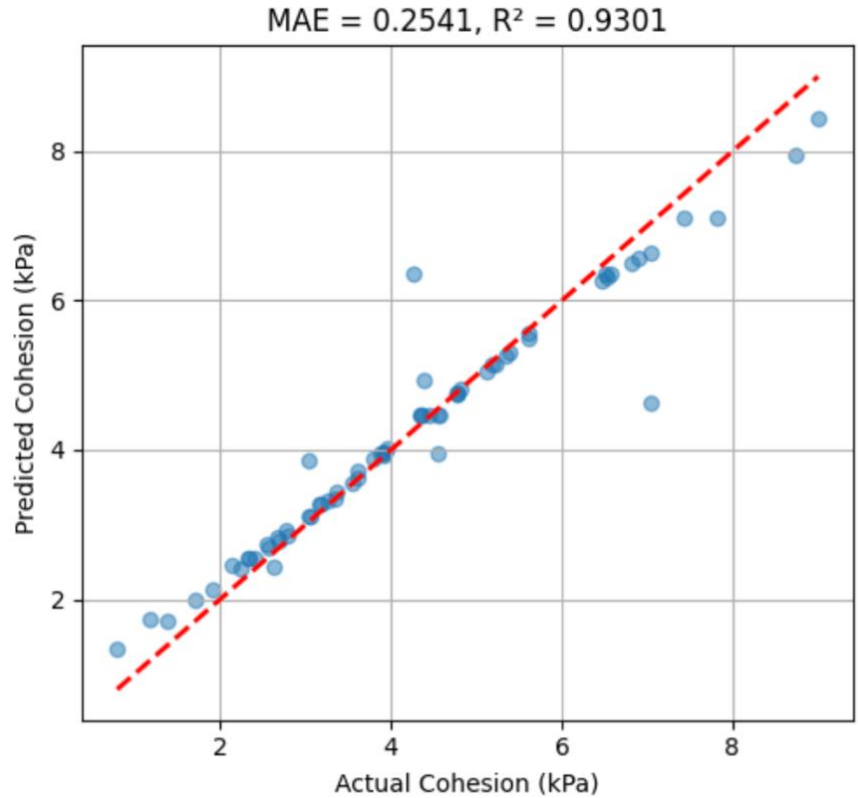

**Fig. 10** Model Performance for Cohesion Prediction: Predicted vs. Actual Values (kPa)
(MAE = 0.2541, $R^2$ = 0.9301)

The figures provided offer a detailed illustration of the AI model's performance in predicting friction angle and cohesion (Figs. 9 and 10). Fig. 9 depicts the relationship between the predicted friction angle values and the actual values. The Mean Absolute Error (MAE) of 1.0377 degrees indicates a relatively small average difference between the predicted and actual friction angles, suggesting that the model's predictions are quite accurate. The R-squared value of 0.9301 further supports this, as it signifies that 93.01% of the variance in the actual friction angle values is explained by the model's predictions, indicating a strong fit. Similarly, Fig. 10 illustrates the model's performance in predicting cohesion. The MAE of 0.2541 kPa is notably low, reinforcing the model's high level of accuracy in predicting cohesion values. The same R-squared value of 0.9301 indicates that the model's predictions closely match the actual cohesion values, with 93.01% of the variance in the actual cohesion values being explained by the model's predictions. These results highlight the robustness and reliability of the AI model in both friction angle and cohesion predictions. The low MAE values and high R-squared values demonstrate the model's ability to accurately capture the underlying patterns in the data, making it a valuable tool for applications that require precise predictive analytics. The consistent performance across both metrics underscores the model's effectiveness and its potential for practical implementation in various fields, such as civil engineering and geotechnical analysis, where accurate predictions of friction angle and cohesion are critical for decision-making and risk assessment.

## 5. Ablation study

The ablation study presented in Table 1 evaluates the performance of various machine learning models in predicting friction angle and cohesion, using the Mean Absolute Percentage Error (MAPE) as the primary evaluation metric. The results demonstrate that the proposed MLP model (MLP [64, 1000, 200, 8]) significantly outperforms both gradient-boosting algorithms (XGBoost and CatBoost) and alternative MLP configurations. Specifically, the proposed model achieved the lowest MAPE values for both friction angle (7.42%) and cohesion (14.96%), indicating superior predictive accuracy. The gradient-boosting algorithms showed comparable performance, with CatBoost achieving a slightly lower MAPE for cohesion (19.38%) compared to XGBoost (21.38%), likely due to its native handling of categorical features and ordered boosting mechanism. Among the MLP configurations, the deeper architecture (MLP [64, 5000, 1000, 200, 8]) underperformed for cohesion (20.19% MAPE), likely due to overfitting from excessive model complexity, while the smaller MLP (MLP [20, 200, 200, 8, 1]) achieved a lower MAPE for cohesion (15.24%) but higher friction angle MAPE (12.28%). The proposed MLP's optimized architecture balances model complexity and generalization, avoiding overfitting while effectively capturing the underlying patterns in the data. These findings highlight the importance of tailored model architecture in geotechnical predictive analytics, where accurate predictions of friction angle and cohesion are critical for applications such as slope stability analysis and foundation design in civil engineering. The proposed model's low MAPE values underscore its reliability and potential for practical implementation, reducing the need for extensive hyperparameter tuning compared to gradient-boosting algorithms. Future work could explore further optimization techniques (e.g., dropout, batch normalization) or hybrid models combining the strengths of gradient-boosting and neural networks to enhance predictive performance.

Table 1 The ablation study of model

| Model | Friction angle MAPE (%) | Cohesion MAPE (%) |
| --- | --- | --- |
| XGBoost | 15.11 | 21.38 |
| CatBoost | 15.12 | 19.38 |
| MLP [20,200,200,8,1] | 12.28 | 15.24 |
| MLP [64, 5000, 1000, 200, 8] | 8.85 | 20.19 |
| **Proposed model MLP [64, 1000, 200, 8]** | 7.42 | 14.96 |

## 6. Explainable Model

In this section, we employ SHAP (SHapley Additive exPlanations) [40] analysis to interpret the model's predictions by examining both global and local feature contributions. The global SHAP analysis provides an overview of how each feature influences the model's overall prediction performance. Specifically, it quantifies the average contribution of each feature across the entire dataset, enabling the identification of key features that drive the model's decision-making process. This global perspective is crucial for understanding the relative importance of features and their impact on the model's predictive accuracy. Local SHAP values, on the other hand, offer instance-specific explanations by detailing the contribution of each feature to individual predictions. This allows for a granular understanding of how specific feature values affect the model's output for particular data points. For example, in the context of friction angle and cohesion prediction, local SHAP values can highlight which features (e.g., soil type, particle size distribution) have the most significant impact on the predicted values for specific instances.

This detailed insight is invaluable for diagnosing model behavior and identifying potential sources of prediction errors.

SHAP analysis is based on Shapley values from cooperative game theory, which provide a fair and mathematically robust way to attribute the contribution of each feature to the model's prediction. The SHAP framework ensures that the sum of the feature contributions equals the difference between the model's prediction and the average prediction, satisfying properties such as local accuracy, missingness, and consistency. By leveraging SHAP, we can enhance the interpretability of complex machine learning models, making them more transparent and trustworthy for applications in civil engineering and geotechnical analysis. The global and local SHAP analyses together provide a comprehensive understanding of the model's behavior, supporting both model optimization and practical decision-making.

SHAP (SHapley Additive exPlanations) analysis was employed to interpret the model's predictions by quantifying the contribution of each feature to both global and local decision-making processes. The fundamental equation for calculating SHAP values for a specific feature i is given by:

$$\phi_i(v, x) = \sum_{S \subseteq N \setminus \{i\}} \frac{|S|!(|N|-|S|-1)!}{|N|!} [v(S \cup \{i\}) - v(S)] \qquad (25)$$

Here, $\phi_i$ represents the SHAP value for feature, $N$ is the complete set of features, $S$ represents any subset of features excluding i, $v$ is the prediction function and $x$ is the specific being explained. The term $\Delta_i(S) = v(S \cup \{i\}) - v(S)$ denotes the marginal contribution of feature $i$ to the coalition $S$, while the coalition weight:

$$w(|S|) = \frac{|S|!(|N|-|S|-1)!}{|N|!} \qquad (26)$$

It ensures fair attribution by accounting for the size of the coalition relative to the total feature set. For neural network models, the prediction function $v(S)$ is defined as the expected prediction given the subset $S$ of features:

$$v(S) = E[f(x)|x_S] \qquad (27)$$

where $x_S$ denotes the instance with only features in $S$ activated, and the remaining features are masked or set to baseline values. This allows SHAP to decompose the model's prediction into an additive feature attribution:

$$f(x) = \phi_0 + \sum_{i=1}^{M} \phi_i \qquad (28)$$

For a friction angle prediction model:

$$\text{Friction Angle}(x) = E[\text{Friction Angle}] + \sum_{i=1}^{M} \phi_i \qquad (29)$$

Given the computational complexity of exact SHAP value calculation, the KernelExplainer approximates SHAP values using a weighted linear regression framework:

$$\phi_i \approx \sum_{z' \in Z} \frac{|Z|}{|Z'||Z|} \left[ \hat{f}(h_x(z')) - \hat{f}\left(h_{x \setminus i}(z')\right) \right] \tag{30}$$

Here, $Z$ is a background dataset, $h_x$ maps simplified inputs (binary feature presence/absence) to the original feature space, and $\hat{f}$ is the trained model. The weights are derived from the proximity of background samples to the instance $x$, ensuring that local patterns are prioritized. The regression weights for the linear approximation are given by:

$$w_i = \frac{|N|-1}{\binom{|N|-1}{|z_i|}|z_i|(|N|-|z_i|)} \tag{31}$$

where $|z_i|$ is the number of non-zero elements in each sample, balancing the influence of feature coalitions.

SHAP values adhere to critical properties ensuring robustness:

1. **Local Accuracy**: The sum of SHAP values plus the base value exactly reconstructs the model's prediction:

$$f(x) = g(x') = \phi_0 + \sum_{i=1}^{M} \phi_i \tag{32}$$

2. **Missingness**: If a feature's value matches the baseline ($x_i = x'_i$), its SHAP value is zero ($\phi_i$=0).
3. **Consistency**: If a feature's marginal contribution increases, its SHAP value does not decrease.

Sequential Attribution in waterfall plots visually decomposes predictions stepwise:

$$\text{Base} \xrightarrow{+\phi_1} \text{Step 1} \xrightarrow{+\phi_2} \text{Step 2} \ldots \xrightarrow{+\phi_M} \text{Final Prediction} \tag{33}$$

This progression highlights how individual features incrementally modify the base prediction, providing transparency for stakeholders. By leveraging SHAP's mathematical framework, which ensures fair attribution, additivity, consistency, and local accuracy, we enhance the interpretability of complex models while maintaining alignment with theoretical guarantees from cooperative game theory. This approach is particularly valuable for geotechnical applications, where model transparency is critical for trust and validation.

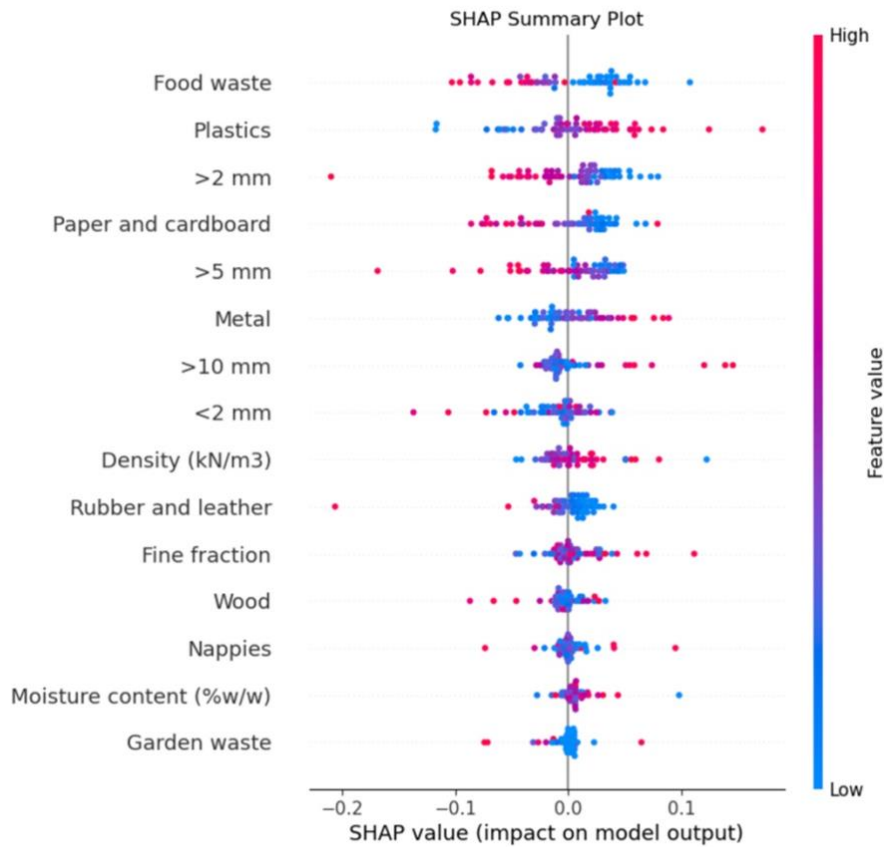

**Fig. 11** Global SHAP value for friction angle

The SHAP (SHapley Additive exPlanations) summary plot (Fig. 11) quantitatively illustrates the relative importance and directional influence of various features on Municipal Solid Waste (MSW) friction angle predictions. The analysis reveals that compositional elements and physical properties exhibit distinct patterns of influence on the model output. Food waste demonstrates predominantly negative SHAP values, indicating its inverse relationship with friction angle, which can be attributed to its high compressibility and reduced interparticle friction characteristics [41]. Conversely, plastic content shows a positive correlation with friction angle, potentially due to internal reinforcement effects at the low confining pressures employed in this study [42], rather than the friction-reducing behavior observed at higher confining pressures in previous research [23]. Particle size distribution emerges as a critical determinant, with fine fractions (<2 mm) generally corresponding to decreased friction angles due to reduced particle interlocking, while coarser fractions (>2 mm, 5-10 mm) contribute to enhanced shear resistance through improved mechanical interlocking [43]. The analysis also indicates significant contributions from bulk physical properties, where density ($kN/m^3$) exhibits a positive correlation with friction angle, consistent with improved granular interlocking at higher densities [12]. Moisture content (%w/w) demonstrates a negative influence, likely due to its lubricating effect on particle interfaces [44].These findings provide quantitative validation of established geotechnical principles regarding MSW mechanical behavior, while offering novel insights into the relative magnitude of each parameter's influence. The SHAP analysis framework enables systematic evaluation of feature importance, facilitating evidence-based optimization of waste management strategies for enhanced structural stability. This comprehensive understanding of compositional and physical property influences supports more informed decision-making in MSW geotechnical applications.

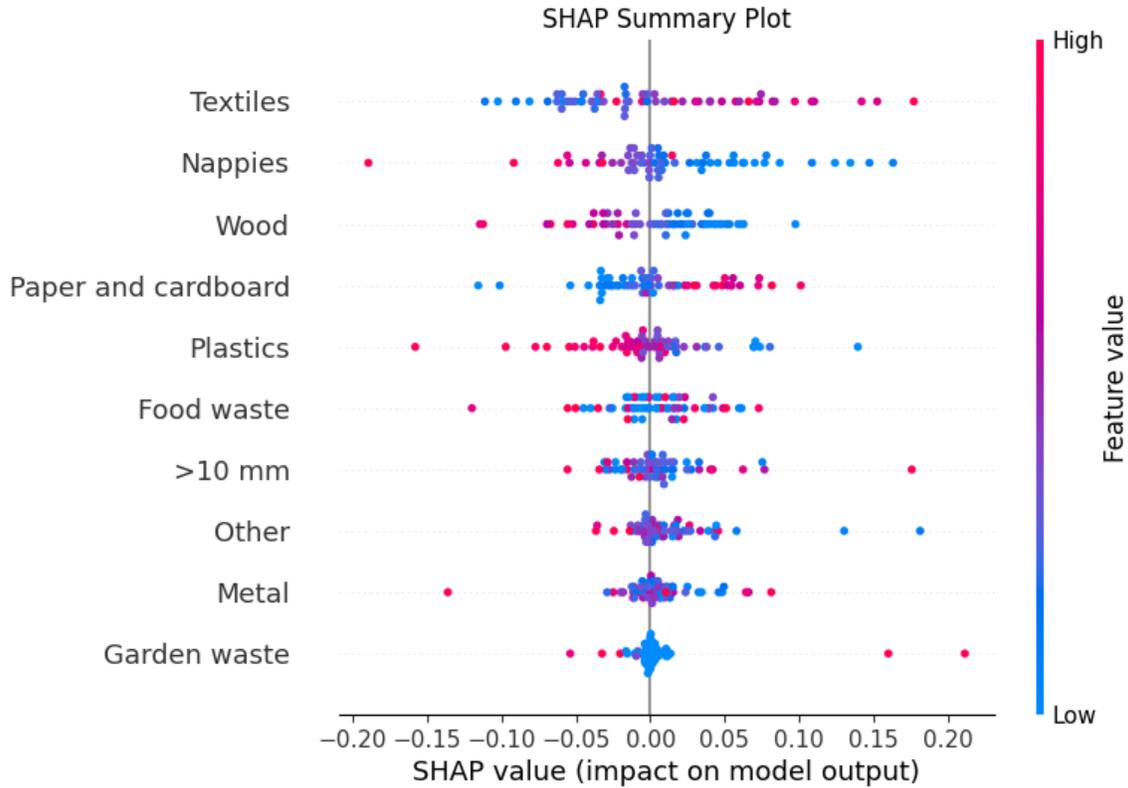

**Fig. 12** Global SHAP value for cohesion of MSW

The SHAP analysis presented in Fig. 12 reveals complex relationships between Municipal Solid Waste (MSW) compositional elements and cohesion through quantitative impact assessment. Textiles demonstrate widely distributed SHAP values around zero, reflecting their dual nature in the waste matrix - potentially providing fiber reinforcement when dry while possibly compromising stability under high moisture conditions [11]. Nappies show predominantly negative effects on cohesion, attributed to their high compressibility and the presence of super-absorbent polymers that create zones of concentrated moisture, potentially weakening structural integrity [4, 45]. Paper and cardboard materials exhibit a slight positive trend in their SHAP values, explained by their ability to form dense, interconnected layers under compression and create mechanical interlocking through their fibrous nature [46]. Their partial decomposition into finer particles can enhance matrix density by filling voids, contributing to improved cohesive strength [43]. This beneficial effect underscores the potential value of maintaining appropriate proportions of these materials in waste management strategies.

Food waste displays variable effects on cohesion, with SHAP values distributed across both positive and negative ranges, reflecting its complex behavior influenced by multiple factors. This variability can be attributed to fresh food waste providing temporary cohesion through natural adhesion, while decomposed material may create unstable zones [47]. Additionally, moisture content fluctuations affect effective stress distributions, leading to dynamic changes in cohesive properties over time [35]. Garden waste and metal components show neutral to slightly negative impacts, with tightly clustered SHAP values near zero, suggesting consistent but limited influence on cohesion. Garden waste's balanced composition of fibrous and organic materials neither significantly enhances nor detracts from overall cohesion, while metals' rigid nature and smooth surfaces may create matrix discontinuities without contributing substantially to binding. These findings emphasize the dynamic nature of waste composition effects on MSW cohesion and underscore the importance of considering both immediate and long-term material property changes in waste management strategies. The complex interactions revealed by the SHAP analysis support the development of more nuanced approaches to waste placement and composition management for enhanced landfill stability. This understanding particularly highlights the potential benefits of maintaining optimal proportions of paper and cardboard materials while carefully managing the distribution of materials with variable or negative impacts on structural integrity.

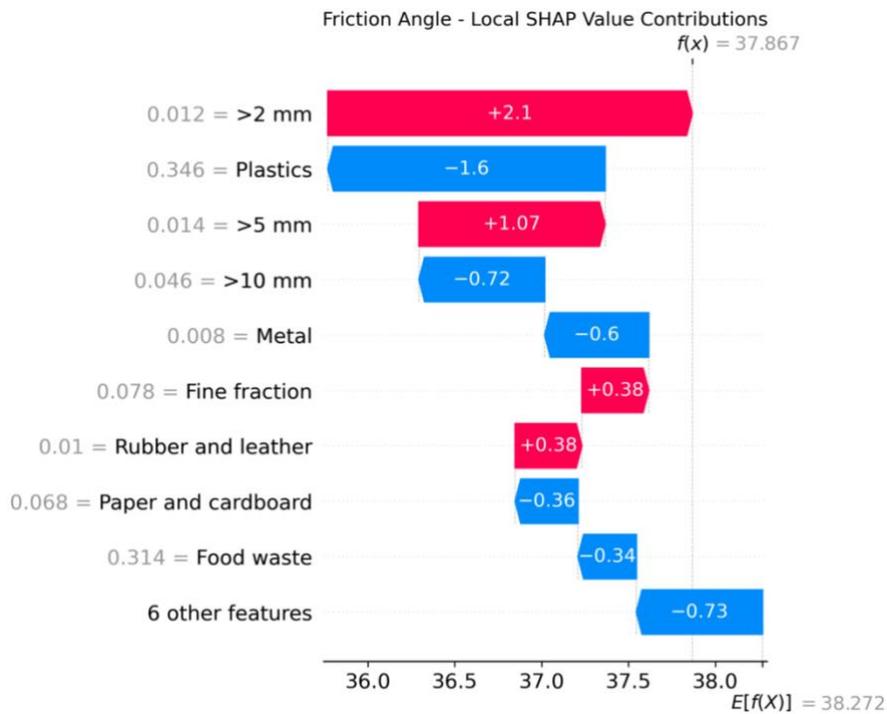

Fig. 13 Local SHAP value for friction angle

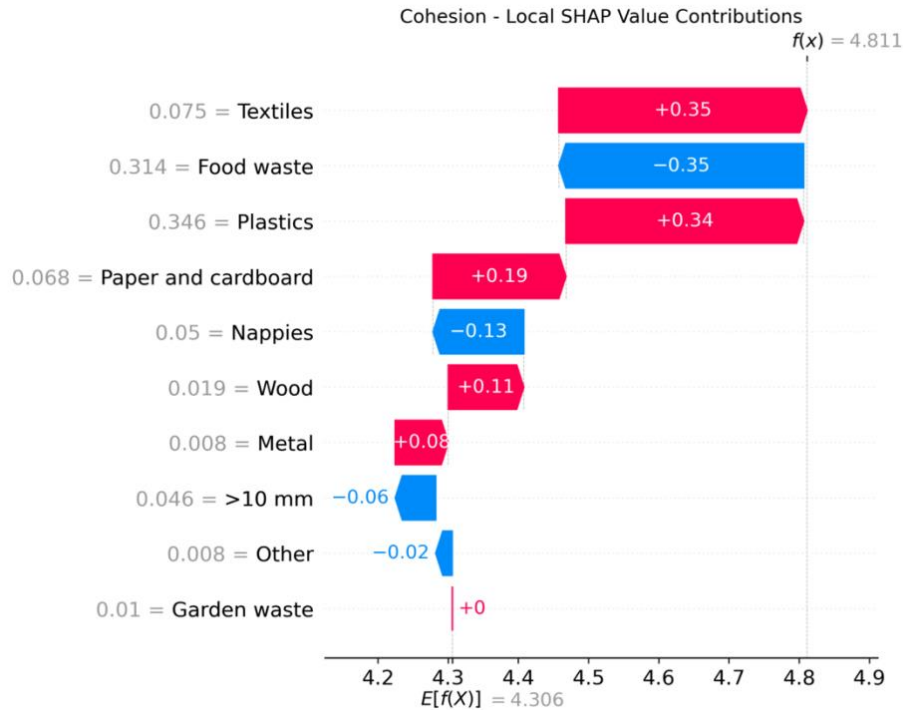

Fig. 14 Local SHAP value for cohesion

Figs. 13 and 14 reveal distinctive mechanisms governing the shear strength parameters of municipal solid waste through SHAP value analysis, demonstrating complex interactions between physical and material properties. In Fig. 12, friction angle predictions show particle size distribution as the dominant factor, with particles >2mm exhibiting the strongest positive contribution (+2.1°), followed by a decreasing trend for larger sizes (>5mm: +1.07°, >10mm: -0.72°). This pattern aligns with fundamental soil mechanics principles, where optimal particle size ranges enhance interlocking and force transmission through the waste matrix [48]. The negative impact of plastics (-1.6°) on friction angle can be attributed to their inherent surface properties and potential shape effects [23], while fine fractions contribute positively (+0.38°) through void-filling mechanisms and increased particle contacts [12].

Fig. 14 demonstrates that cohesion exhibits stronger dependence on material type rather than physical dimensions. Textiles and plastics demonstrate the highest positive contributions to cohesion (+0.35 and +0.34, respectively), likely due to fiber reinforcement effects and material entanglement in the waste matrix [49]. The notable negative impact of food waste (-0.35) on cohesion suggests the influence of moisture content and degradation processes on inter-particle bonding [47]. These opposing trends between friction and cohesion for similar materials (e.g., plastics) highlight the complex nature of waste mechanical behavior and the importance of considering multiple strength parameters in stability assessments.

The comparative analysis of SHAP values between Figs. 12 and 13 indicates that friction angle exhibits greater sensitivity to composition changes, with larger magnitude variations (-1.6° to +2.1°) compared to cohesion (-0.35 to +0.35). This finding has significant implications for waste management practices, suggesting that particle size optimization could provide more substantial improvements in frictional strength than in cohesive behavior. The identified relationships provide valuable insights for engineering applications, including material acceptance criteria, waste placement procedures, and stability analyses. Furthermore, the SHAP analysis demonstrates the utility of machine learning interpretability techniques in understanding complex geotechnical systems, bridging the gap between empirical observations and mechanistic understanding in waste mechanics. These findings from Figs. 13 and 14 contribute to both theoretical frameworks and practical applications in landfill engineering, offering quantitative guidance for optimizing waste composition and predicting mechanical behavior. The study underscores the importance of considering material-specific contributions to shear strength parameters, particularly in heterogeneous waste systems where traditional analytical approaches may be limited. The results suggest that strategic management of waste composition and size distribution could enhance overall mechanical stability, while accounting for the distinct mechanisms governing friction and cohesion in waste materials.

## 7. Application

The implementation of a machine learning-based web application for waste properties prediction, as illustrated in Fig. 15, demonstrates a sophisticated integration of computational methods with geotechnical engineering principles. [https://huggingface.co/spaces/Sompote/MSW_Shear]. The application's robust framework incorporates seventeen critical parameters, carefully selected to characterize waste material properties comprehensively. These parameters span particle size distributions ranging from <2mm to 10mm, physical properties including density (measured at 7.23 kN/m$^3$) and fine fraction (0.08), and detailed composition parameters covering materials from food waste (0.31) to plastics (0.35) and textiles (0.08). As shown in Fig. 14, the interface is organized into two main sections: Input Parameters and Prediction Results, where users can either upload Excel files (XLSX, XLS format, limited to 200MB) or manually input parameters. The model's output validation is evidenced by its predictions falling within expected ranges for municipal solid waste, producing a friction angle of 37.87° and cohesion of 4.81 kPa.

The application's sophistication is further enhanced by its integration of SHAP (SHapley Additive exPlanations) analysis, which provides transparent interpretation of the prediction mechanisms. The SHAP analysis visualizations, displayed in the lower section of Fig. 15, reveal important relationships between waste components and mechanical properties, showing how smaller particles and fibrous materials positively influence friction angle, while textiles and food waste contribute significantly to cohesion. This aligns with established geotechnical principles and provides crucial validation of the model's learning patterns. The technical implementation offers remarkable flexibility through its dual input methodology, accommodating both direct manual entry for individual analyses and Excel file uploads for batch processing. This design choice reflects a deep understanding of varying user needs in different engineering scenarios. The real-time calculation capability, coupled with immediate visual interpretation through SHAP analysis, transforms complex waste mechanics analysis into an accessible and efficient process. The application's practical value extends beyond mere prediction, offering a comprehensive decision support system for engineering practice. Its ability to provide immediate assessment while maintaining transparency through detailed parameter contribution analysis represents a significant advancement over traditional empirical methods.

The system effectively addresses the complex nature of waste mechanics by incorporating material interactions while maintaining consistency with established geotechnical principles. This balance between sophisticated analysis and practical usability makes the application particularly valuable in scenarios where understanding the reasoning behind predictions is crucial for engineering decisions. Moreover, the application's ability to handle multivariate analysis while providing clear visualization of parameter influences addresses a critical need in waste mechanics engineering. The SHAP visualizations, as demonstrated in Fig. 15, effectively bridge the gap between complex machine learning operations and practical engineering requirements, enabling engineers to verify predictions against physical principles and identify critical parameters influencing waste mechanical behavior. This comprehensive approach demonstrates how artificial intelligence can be effectively applied to enhance traditional engineering practices while maintaining the transparency and reliability required for professional applications.

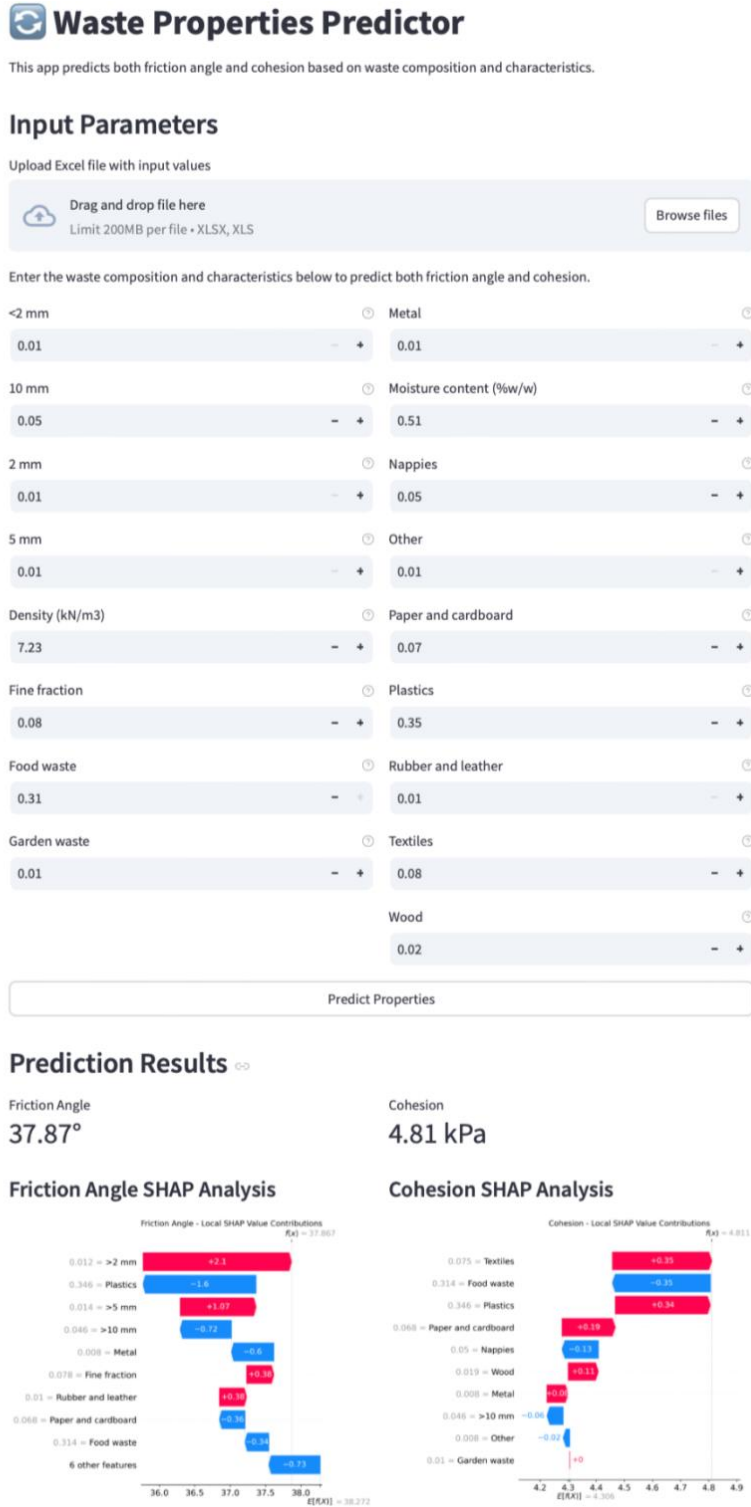

**Fig. 15** Interactive web-based application for predicting municipal solid waste shear strength parameters (friction angle and cohesion) with integrated SHAP (SHapley Additive exPlanations) value computation. [https://huggingface.co/spaces/Sompote/MSW_Shear]

## 8. Discussion

The explainable AI framework developed for predicting MSW shear strength parameters demonstrates both promising capabilities and notable limitations that warrant discussion. Through 10-fold cross-validation, the model achieved consistent performance across different data subsets, with average Mean Absolute Error (MAE) values of 5.1±1.4 degrees for friction angle and 0.5±0.3 kPa for cohesion, demonstrating robust generalization despite the limited dataset size. The narrow standard deviations in these metrics suggest stable predictive capabilities across different waste compositions. The model's superior performance compared to traditional gradient boosting approaches (XGBoost: 15.11% MAPE, CatBoost: 15.12% MAPE) for friction angle prediction, achieving 7.42% MAPE, validates the effectiveness of the proposed deep learning architecture.

The SHAP analysis revealed significant insights into feature importance, with food waste, plastics, and particle size fractions emerging as key determinants of friction angle prediction. This aligns with established geotechnical understanding, where these components significantly influence MSW mechanical behavior. Particularly noteworthy is the model's ability to capture the complex relationships between waste composition and shear strength parameters, as evidenced by the non-linear effects observed in the SHAP value distributions. The local interpretability provided by SHAP waterfall plots offers practical value for engineers, enabling detailed understanding of how specific waste compositions influence stability predictions.

However, several limitations must be acknowledged. While the model achieves strong predictive performance, the relatively small dataset size (66 samples) from a single dumpsite location presents inherent limitations for model generalization. The current architecture, though effective for the available data, may not fully capture the complex temporal dependencies inherent in waste degradation processes. This limitation is particularly relevant given that MSW properties evolve significantly over time through biochemical decomposition and physical settling processes.

Looking forward, several promising research directions emerge from these limitations. The model architecture and pre-trained weights could serve as a foundation for transfer learning applications to other dumpsites, particularly those with similar waste composition profiles in Southeast Asia. This approach could help address the challenge of limited data availability at new sites while leveraging the knowledge gained from the current dataset. The development of larger, more diverse datasets incorporating samples from multiple landfills across different geographical regions would enhance model robustness and generalizability. Integration of temporal monitoring data could better capture degradation effects on shear strength parameters, while standardized waste characterization protocols would improve data consistency.

From an architectural perspective, investigating hybrid models that combine CNN-LSTM networks could better capture spatial-temporal patterns in waste behavior. The implementation of multi-task learning approaches might enable simultaneous prediction of multiple mechanical properties, potentially improving overall model utility for practical applications. The experimental constraints of direct shear testing under specific normal stress conditions (10.00 to 20.00 kPa) and shallow burial depths (1.23 to 4.77 m) may not comprehensively represent the full range of conditions encountered in operational landfills.

The practical implementation of this framework could be significantly enhanced through integration with real-time monitoring systems and user-friendly interfaces for engineering practitioners. The potential integration with remote sensing technologies and IoT-based systems presents an opportunity for continuous data collection and model updating, potentially leading to more dynamic and responsive prediction capabilities. These advancements could bridge the current gap between laboratory-based measurements and field-scale applications, ultimately improving our ability to assess and manage landfill stability.

This research represents a significant step toward integrating advanced AI techniques with geotechnical engineering practice, though continued development is needed to address the identified limitations and expand the framework's practical utility. The balance between model sophistication and interpretability remains a key consideration, particularly in geotechnical applications where engineering judgment plays a crucial role in decision-making processes. Future work focusing on transfer learning and domain adaptation could extend the model's applicability to diverse waste management contexts while maintaining its core predictive capabilities.

## 9. Conclusion

This research presents a novel explainable artificial intelligence framework for predicting shear strength parameters of municipal solid waste (MSW) across diverse compositional profiles. The study's key contributions and findings can be summarized as follows:

The proposed multi-layer perceptron architecture (MLP [64, 1000, 200, 8]) demonstrated superior predictive performance compared to traditional gradient boosting methods, achieving mean absolute percentage errors of 7.42% and 14.96% for friction angle and cohesion predictions, respectively. The model's robust generalization capability was validated through comprehensive 10-fold cross-validation, yielding consistent performance metrics with MAE values of 5.1±1.4 degrees for friction angle and 0.5±0.3 kPa for cohesion.

Integration of SHAP (SHapley Additive exPlanations) analysis provided crucial insights into the relationship between waste composition and mechanical properties. The analysis revealed that fibrous materials and particle size distribution significantly influence shear strength parameters, with food waste and plastics showing notable but non-linear effects. This quantitative understanding of feature importance aligns with established geotechnical principles while offering new perspectives on component interactions.

The framework's explainability component successfully bridges the gap between advanced machine learning techniques and practical engineering applications. Through transparent feature attribution and local interpretability, the model provides engineers with actionable insights for waste management decisions while maintaining scientific rigor. The implementation of both global and local SHAP analyses enables multi-scale understanding of waste behavior, from overall compositional trends to specific instance predictions.

The study's methodology, combining large-scale direct shear testing with sophisticated AI modeling, establishes a foundation for future research in MSW characterization. The framework's architecture and pre-trained weights offer potential for transfer learning applications to other dumpsites, particularly in regions with similar waste management contexts. This adaptability, coupled with the model's strong predictive performance, provides a practical tool for rapid assessment of MSW mechanical properties.

While limitations exist, primarily regarding dataset size and geographical representation, this research demonstrates the feasibility and value of explainable AI approaches in geotechnical engineering. The framework's success in maintaining both predictive accuracy and interpretability suggests promising applications in landfill design, stability assessment, and waste management optimization. Future work focusing on expanded datasets, temporal monitoring, and integration with field measurements will further enhance the model's utility in practical engineering scenarios. This study ultimately contributes to the growing intersection of artificial intelligence and geotechnical engineering, offering a reliable, interpretable, and practical approach to predicting MSW mechanical behavior. The framework's success in balancing sophisticated machine learning techniques with engineering practicality provides a template for future developments in this critical field.

## Statements and Declarations
### Funding
The study was financially supported for laboratory testing and data acquisition by Petchra Pra Jom Klao Doctoral Scholarship No 38/2565 from King Mongkut's University of Technology Thonburi, The Joint Graduate School of Energy and Environment (JGSEE), King Mongkut's University of Technology Thonburi (KMUTT), and the Center of Excellence on Energy Technology and Environment (CEE), PERDO, and Ministry of Higher Education, Science, Research and Innovation (MHESI), Thailand.

### Acknowledgements
We would like to express our gratitude to the AI Research Group, for providing software support throughout this study and to Eastern Energy Plus Company Limited (EEP) for supporting permission to conduct surveys at the solid waste disposal site.

### Declarations
The authors have no competing interests to declare that are relevant to the content of this article.

## References
1. Liu C, Shi J, Lv Y, Shao G (2021) A modified stability analysis method of landfills dependent on gas pressure. Waste Manag Res 39:784–794. https://doi.org/10.1177/0734242X20944479

2. Maalouf A, Mavropoulos A (2023) Re-assessing global municipal solid waste generation. Waste Manag Res 41:936–947. https://doi.org/10.1177/0734242X221074116

3. Zhang Z, Wang Y, Fang Y, et al (2020) Global study on slope instability modes based on 62 municipal solid waste landfills. Waste Manag Res 38:1389–1404. https://doi.org/10.1177/0734242X20953486

4. Huang M, Zhang Z, Zhu B, et al (2023) Effects of moisture content and landfill age on the shear strength properties of municipal solid waste in Xi'an, China. Environmental Science and Pollution Research 30:65011–65025. https://doi.org/10.1007/s11356-023-26905-6

Explainable Artificial Intelligence Model for Evaluating Shear Strength Parameters of Municipal Solid Waste Across Diverse Compositional Profiles     31

5.  Stark T, Huvaj N, Li G (2008) Shear strength of municipal solid waste for stability analyses. Environmental Geology 57:1911–1923. https://doi.org/10.1007/s00254-008-1480-0

6.  Keramati M, Shahedifar M, Aminfar MH, Alagipuor H (2020) Evaluation the Shear Strength Behavior of aged MSW using Large Scale In Situ Direct Shear Test, a case of Tabriz Landfill. International Journal of Civil Engineering 18:717–733. https://doi.org/10.1007/s40999-020-00499-3

7.  Kolapo P, Oniyide G, Said K, et al (2022) An Overview of Slope Failure in Mining Operations

8.  Youwai S, Detcheewa S (2025) Predicting rapid impact compaction of soil using a parallel transformer and long short-term memory architecture for sequential soil profile encoding. Engineering Applications of Artificial Intelligence 139:109664. https://doi.org/10.1016/j.engappai.2024.109664

9.  Youwai S, Makam P (2024) CTrPile: A Computer Vision and Transformer Approach for Pile Capacity Estimation from Dynamic Pile Load Test. In: Proceeding of 2024 IEEE Conference on Artificial Intelligence (IEEE CAI 2024). Singapore

10. Norberto A, Medeiros R, Corrêa C, et al (2024) Laboratory tests for evaluation of shear strength and tensile effect generated by fibers present in Muribeca's landfills of municipal solid waste. Soils and Rocks 47:e2024008622. https://doi.org/10.28927/SR.2024.008622

11. Pulat HF, Yukselen-Aksoy Y (2016) Composition Effects on the Shear Strength Behavior of Synthetic Municipal Solid Wastes. In: Geo-Chicago 2016. American Society of Civil Engineers, Chicago, Illinois, pp 101–110

12. Bray J, Zekkos D, Kavazanjian E, et al (2009) Shear Strength of Municipal Solid Waste. Journal of Geotechnical and Geoenvironmental Engineering - J GEOTECH GEOENVIRON ENG 135:. https://doi.org/10.1061/(ASCE)GT.1943-5606.0000063

13. Chen D, Chen Y, Deng Y, et al (2024) Mechanical properties of municipal solid waste under different stress paths: Effects of plastic content and particle gradation. Waste Management 185:43–54. https://doi.org/10.1016/j.wasman.2024.05.041

14. Dixon N, Jones DRV (2005) Engineering properties of municipal solid waste. Geotextiles and Geomembranes 23:205–233. https://doi.org/10.1016/j.geotexmem.2004.11.002

15. Youwai S, Wongsala K (2024) Predicting the Friction Angle of Bangkok Sand Using State Parameter and Neural Network. Geotech Geol Eng. https://doi.org/10.1007/s10706-024-02873-7

16. Chen X, Ma C, Ren Y, et al (2023) Explainable artificial intelligence in finance: A bibliometric review. Finance Research Letters

17. Lai T (2024) Interpretable Medical Imagery Diagnosis with Self-Attentive Transformers: A Review of Explainable AI for Health Care. BioMedInformatics

18. Abdollahi A, Li D, Deng J, Amini A (2024) An explainable artificial-intelligence-aided safety factor prediction of road embankments. Engineering Applications of Artificial Intelligence 136:108854. https://doi.org/10.1016/j.engappai.2024.108854

19. Hsiao C-H, Kumar K, Rathje EM (2024) Explainable AI models for predicting liquefaction-induced lateral spreading. Front Built Environ 10:1387953. https://doi.org/10.3389/fbuil.2024.1387953

20. Liu F, Liu W, Li A, Cheng JCP (2024) Geotechnical risk modeling using an explainable transfer learning model incorporating physical guidance. Engineering Applications of Artificial Intelligence 137:109127. https://doi.org/10.1016/j.engappai.2024.109127

21. Gordan B, Jahed Armaghani D, Hajihassani M, Monjezi M (2016) Prediction of seismic slope stability through combination of particle swarm optimization and neural network. Engineering with Computers 32:85–97. https://doi.org/10.1007/s00366-015-0400-7

22. Boonsakul P, Suanburi D, Towprayoon S, et al (2024) Advanced ERT techniques for methane potential evaluation in controlled dump sites: A forward modeling approach. Results in Engineering 24:103256. https://doi.org/10.1016/j.rineng.2024.103256

23. Singh V, Uchimura T (2023) Effect of Material Composition on Geotechnical Properties—Study on Synthetic Municipal Solid Waste. Geotechnics 3:397–415. https://doi.org/10.3390/geotechnics3020023

24. Abdallah M, Abu Talib M, Feroz S, et al (2020) Artificial intelligence applications in solid waste management: A systematic research review. Waste Management 109:231–246. https://doi.org/10.1016/j.wasman.2020.04.057

25. Hsiao C-H, Chen AY, Ge L, Yeh F-H (2022) Performance of artificial neural network and convolutional neural network on slope failure prediction using data from the random finite element method. Acta Geotech 17:5801–5811. https://doi.org/10.1007/s11440-022-01520-w

26. Mu'azu MA (2023) Enhancing Slope Stability Prediction Using Fuzzy and Neural Frameworks Optimized by Metaheuristic Science. Math Geosci 55:263–285. https://doi.org/10.1007/s11004-022-10029-7

27. Muttaraid A, Towprayoon S, Chiemchaisri C, et al (2024) Optimizing landfill mining operations: a comparative analysis of material flow and cost–benefit assessment for medium- and large-scale sites. J Mater Cycles Waste Manag 26:830–844. https://doi.org/10.1007/s10163-023-01853-y

28. Shi J, Shu S, Ai Y, et al (2021) Effect of elevated temperature on solid waste shear strength and landfill slope stability. Waste Manag Res 39:351–359. https://doi.org/10.1177/0734242X20958065

29. Siddiqua A, Hahladakis JN, Al-Attiya WAKA (2022) An overview of the environmental pollution and health effects associated with waste landfilling and open dumping. Environ Sci Pollut Res 29:58514–58536. https://doi.org/10.1007/s11356-022-21578-z

30. Sutthasil N, Chiemchaisri C, Chiemchaisri W, et al (2019) The effectiveness of passive gas ventilation on methane emission reduction in a semi-aerobic test cell operated in the tropics. Waste Management 87:954–964. https://doi.org/10.1016/j.wasman.2018.12.013

31. Wangyao K, Sutthasil N, Chiemchaisri C (2021) Methane and nitrous oxide emissions from shallow windrow piles for biostabilisation of municipal solid waste. Journal of the Air & Waste Management Association 71:650–660. https://doi.org/10.1080/10962247.2021.1880498

32. Chaulya SK, Prasad GM (2016) Slope Failure Mechanism and Monitoring Techniques. In: Sensing and Monitoring Technologies for Mines and Hazardous Areas. Elsevier, pp 1–86

33. Li X, Peng B, Li J (2023) Constitutive model for stress–Strain responses of municipal solid waste considering fibrous reinforcement. Front Earth Sci 10:1059234. https://doi.org/10.3389/feart.2022.1059234

34. Chiemchaisri C, Charnnok B, Visvanathan C (2010) Recovery of plastic wastes from dumpsite as refuse-derived fuel and its utilization in small gasification system. Bioresource Technology 101:1522–1527. https://doi.org/10.1016/j.biortech.2009.08.061

35. Pecorini I, Iannelli R (2020) Characterization of Excavated Waste of Different Ages in View of Multiple Resource Recovery in Landfill Mining. Sustainability 12:1780. https://doi.org/10.3390/su12051780

36. Chungam B, Vinitnantharat S, Towprayoon S, et al (2021) Evaluation of the potential of refuse-derived fuel recovery in the open dump by resistivity survey prior to mining. J Mater Cycles Waste Manag 23:1320–1330. https://doi.org/10.1007/s10163-021-01207-6

37. Prechthai T, Padmasri M, Visvanathan C (2008) Quality assessment of mined MSW from an open dumpsite for recycling potential. Resources, Conservation and Recycling 53:70–78. https://doi.org/10.1016/j.resconrec.2008.09.002

38. Wang G, Zhao B, Wu B, et al (2023) Intelligent prediction of slope stability based on visual exploratory data analysis of 77 in situ cases. International Journal of Mining Science and Technology 33:47–59. https://doi.org/10.1016/j.ijmst.2022.07.002

39. Huang Y, Hu G (2023) Influence of normal stress on the shear strength of the structural plane considering the size effect. Frontiers in Earth Science 11:. https://doi.org/10.3389/feart.2023.1116302

40. Lundberg SM, Lee S-I (2017) A Unified Approach to Interpreting Model Predictions. In: Guyon I, Luxburg UV, Bengio S, et al (eds) Advances in Neural Information Processing Systems. Curran Associates, Inc., pp 4765–4774

41. Daciolo LVP, Correia NDS, Boscov MEG (2022) Extensive database of MSW shear strength parameters obtained from laboratory direct shear tests: Proposal for data classification. Waste Management 140:245–259. https://doi.org/10.1016/j.wasman.2021.09.015

42. Bareither C, Benson C, Edil T (2012) Effects of Waste Composition and Decomposition on the Shear Strength of Municipal Solid Waste. Journal of Geotechnical and Geoenvironmental Engineering 138:1161–1174. https://doi.org/10.1061/(ASCE)GT.1943-5606.0000702

43. Wang C-H, Fang L, Chang DT-T, et al (2024) Evaluating the strength characteristics of mixtures of municipal solid waste incineration bottom ash and reddish laterite clay for sustainable construction. Heliyon 10:e37780. https://doi.org/10.1016/j.heliyon.2024.e37780

44. Yang R, Xu Z, Chai J, et al (2024) Effect of the spatial changes in engineering properties on municipal solid waste landfill slope stability. Bull Eng Geol Environ 83:462. https://doi.org/10.1007/s10064-024-03952-y

45. Li Y-P, Fan B, Cheng L, et al (2024) Physical and mechanical characterization of solid waste collected from rural areas of China: Experimental study. Process Safety and Environmental Protection 191:873–882. https://doi.org/10.1016/j.psep.2024.09.038

46. Zekkos D, Athanasopoulos GA, Bray JD, et al (2010) Large-scale direct shear testing of municipal solid waste. Waste Management 30:1544–1555. https://doi.org/10.1016/j.wasman.2010.01.024

47. Cho YM, Ko JH, Chi L, Townsend TG (2011) Food waste impact on municipal solid waste angle of internal friction. Waste Management 31:26–32. https://doi.org/10.1016/j.wasman.2010.07.018

48. Li Y (2013) Effects of particle shape and size distribution on the shear strength behavior of composite soils. Bull Eng Geol Environ 72:371–381. https://doi.org/10.1007/s10064-013-0482-7

49. Abreu AES, Vilar OM (2017) Influence of composition and degradation on the shear strength of municipal solid waste. Waste Management 68:263–274. https://doi.org/10.1016/j.wasman.2017.05.038